\documentclass[runningheads]{llncs}
\usepackage{graphicx}
\usepackage{amsmath,amssymb} 
\usepackage{color}
\usepackage[width=122mm,left=12mm,paperwidth=146mm,height=193mm,top=12mm,paperheight=217mm]{geometry}
\usepackage{xspace}
\usepackage{url}
\usepackage{multirow}
\usepackage{makecell}
\usepackage{subcaption}
\usepackage{wrapfig}

\newcommand{\myparagraph}[1]{\smallskip\noindent\textbf{#1.}}
\newcommand{\mysubparagraph}[1]{\smallskip\noindent-- \emph{#1:}}

\begin{document}

\def \ie {\emph{i.e}.\null\xspace}
\def \etal {\emph{et al.}\null\xspace}
\pagestyle{headings}
\mainmatter
\title{Convolutional Networks for Object Category and 3D Pose Estimation from 2D Images} 


\titlerunning{Joint Object Category and 3D Pose Estimation from 2D Images}


	\author{Siddharth Mahendran \and Haider Ali \and Ren{\'e} Vidal \\ {\tt \small \{siddharthm, hali, rvidal\}@jhu.edu}}
	\institute{Center for Imaging Science, Johns Hopkins University}


\maketitle

\begin{abstract}
	Current CNN-based algorithms for recovering the 3D pose of an object in an image assume knowledge about both the object category and its 2D localization in the image. In this paper, we relax one of these constraints and propose to solve the task of joint object category and 3D pose estimation from an image assuming known 2D localization.
	We design a new architecture for this task composed of a feature network that is shared between subtasks, an object categorization network built on top of the feature network, and a collection of category dependent pose regression networks. We also introduce suitable loss functions and a training method for the new architecture.
	Experiments on the challenging PASCAL3D+ dataset show state-of-the-art performance in the joint categorization and pose estimation task. Moreover, our performance on the joint task is comparable to the performance of state-of-the-art methods on the simpler 3D pose estimation with known object category task.
	
\end{abstract}

\section{Introduction}
\label{sec:introduction}
Object pose estimation is the task of estimating the relative rigid transformation between the camera and the object in an image. This is an old and fundamental problem in computer vision and a stepping stone for many other problems such as 3D scene understanding and reconstruction. Recently, this problem has enjoyed renewed interest due to the emergence of applications in autonomous driving and augmented reality, where the ability to reason about objects in 3D is key. 

\begin{wrapfigure}{r}{0.5\linewidth}
	\vspace{-8mm}
	\centering
	\includegraphics[width=\linewidth]{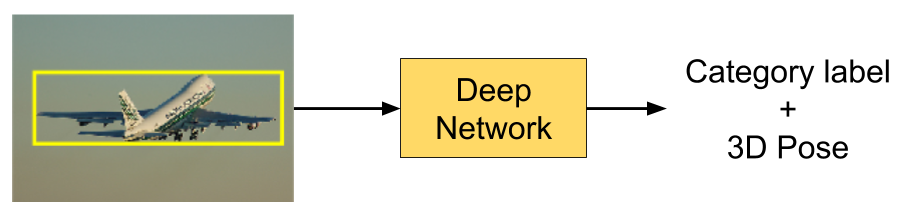}
	\vspace{-7mm}
	\caption{Overview of our problem}
	\label{fig:problem_formulation}
	\vspace{-8mm}
\end{wrapfigure}
As with many computer vision tasks, methods based on convolutional neural networks (CNNs) have been shown to work really well for object pose estimation \cite{Tulsiani:CVPR15,Su:ICCV15,Mousavian:CVPR17,Pavlakos:ICRA17,Wu:ECCV16,Mahendran:ICCVW17}. However, these methods often assume knowledge about the object category and its 2D localization in the image. In this paper, we relax one of these constraints and propose to solve the problem of joint object category and 3D pose estimation from 2D images assuming known localization of the object in the image. 
More specifically, we assume that the bounding box around an object in the image is provided to us by an oracle and we learn a deep network that predicts the category label and 3D pose of the object, as illustrated in Fig.~\ref{fig:problem_formulation}. 

\begin{wrapfigure}{r}{0.67\linewidth}
	\centering
	\includegraphics[width=\linewidth]{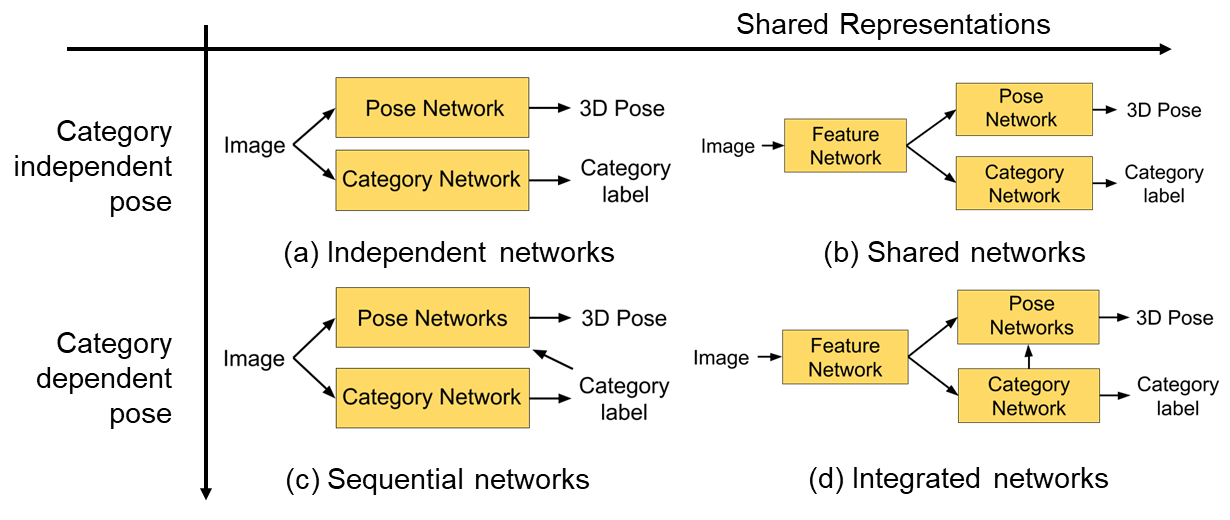}
	\vspace{-7mm}
	\caption{Different network architectures for joint object category and 3D pose estimation}
	\vspace{-7mm}
	\label{fig:category_pose_networks}
\end{wrapfigure}
One approach to object category and 3D pose estimation is to learn independent networks for each task, as illustrated in Fig.~\ref{fig:category_pose_networks}a. 
However, this approach does not exploit the fact that some parts of the representation could be shared across multiple tasks \cite{Kokkinos:CVPR17}. To address this issue, \cite{Elhoseiny:ICML16} designs independent pose and category networks, but each one is built on top of a feature network that is shared by both tasks, as shown in Fig.~\ref{fig:category_pose_networks}b. However, one issue with both independent and shared networks is that they predict pose in a category agnostic manner, which may not always be possible because in some cases it may be difficult to define a universal reference frame (or characteristic pose) for all object categories.\footnote{A natural choice is the center of gravity of the object and the three principal directions (PDs), but even PDs cannot be consistently defined across object categories.} 
To address this issue, we could train a category dependent pose network, i.e., a collection of pose networks (one per object category), each of which takes as input the 2D image and predicts a 3D pose, as shown in Fig.~2c. The final 3D pose predicted by this sequential network is the pose predicted by the network corresponding to the predicted category label. However, as is the case for independent networks, sequential networks do not take advantage of shared representations.

\myparagraph{Paper contributions}
We propose an integrated architecture that provides the best of both worlds by integrating (1) a shared feature representation for both tasks and (2) a category dependent pose network. The proposed architecture consists of a shared feature network, whose output is used as an input to both a category network and a category dependent pose network, as shown in Fig.~2d. The feature network is a residual network learned so that it captures properties of the image that are relevant to both categorization and pose estimation. The category network is also a residual network applied to the output of the feature network. Finally, the category dependent pose network is a collection of fully connected networks (one per object category) that receives the outputs of both the feature and categorization networks as inputs. Since the latter is a class probability vector, it can be used to predict the final pose by fusing pose from individual categories, thereby being potentially more robust to errors in the estimation of the object category. We also devise a new training algorithm for our proposed architecture. Our experiments show that the proposed approach achieves state-of-the-art performance on the challenging Pascal3D+ \cite{Xiang:WACV14} dataset for the joint categorization and pose estimation task; which is comparable to the performance of state-of-the-art methods on the simpler 3D pose estimation with known object category task. We also present an ablative analysis that provides empirical justification for our design choices such as (i) network architecture, (ii) feature representations, (iii) training method, etc. To the best of our knowledge, our work is the first to use residual networks \cite{He:CVPR16} --that have worked very well for the task of image classification-- for 3D pose estimation. We do note that \cite{Hara:arxiv17} also use residual networks but for azimuth or orientation estimation.

\myparagraph{Paper outline} We first review related work in \S\ref{sec:related_work}. We then describe our model for joint object category and 3D pose estimation in \S\ref{sec:model}, including the proposed architecture in \S\ref{sec:network}, loss functions in \S\ref{sec:loss}, and training procedure in \S\ref{sec:network_training}. Finally, we describe our experimental evaluation and analysis in \S\ref{sec:experiments}.

\vspace{-5mm}
\section{Related Work}
\label{sec:related_work}
\vspace{-5mm}
There are two main lines of research that are relevant to our work: (i) 3D pose estimation given object category label and (ii) joint object category and pose estimation. There are many non-deep learning based approaches such as \cite{Saste:ICCVWS11,Hejrati:CVPR14,Aubry:CVPR14,Pepik:CVPR12,Savarese-FeiFei:ICCV07,Bakry:WACV16,Mottaghi:CVPR15,Juranek:ICCV15}, which have designed systems to solve these two tasks. However, due to space constraints, we restrict our discussion to deep learning based approaches. 

\myparagraph{3D Pose estimation given object category label} Current literature for this task can be grouped into two categories: (1) predict 2D keypoints from images and then align 3D models with these keypoints to recover 3D pose and (2) predict 3D pose directly from 2D images. The first category includes methods like Pavlakos \etal \cite{Pavlakos:ICRA17} and Wu \etal \cite{Wu:ECCV16}, which recover a full 6-dimensional pose (azimuth, elevation, camera-tilt, depth, image-translation). Both methods train on images that have been annotated with 2D keypoints that correspond to semantic keypoints on 3D object models. Given a 2D image, they first generate a probabilistic heatmap of 2D keypoints and then recover 3D pose by aligning these 2D keypoints with the 3D keypoints. The second category includes methods like Tulsiani and Malik \cite{Tulsiani:CVPR15}, Su \etal \cite{Su:ICCV15}, Mahendran \etal \cite{Mahendran:ICCVW17}, Mousavian \etal \cite{Mousavian:CVPR17} and Wang \etal \cite{Wang:PCM16}, where they recover the 3D rotation between the object and the camera which corresponds to a 3-dimensional pose (azimuth, elevation, camera-tilt). We also aim to predict a 3-dof pose output in this work. \cite{Tulsiani:CVPR15} and \cite{Su:ICCV15} setup a pose-classification problem by discretizing the euler angles into pose-labels and minimize the  cross-entropy loss during training. \cite{Mousavian:CVPR17} solves a mixed classification-regression problem by returning both pose-label and a residual angle associated with every pose-label, such that the predicted angle is the sum of the center of pose-bin and the corresponding residual angle with the highest confidence. \cite{Mahendran:ICCVW17} and \cite{Wang:PCM16} on the other hand, solve a pose regression problem. While \cite{Wang:PCM16} directly regresses the three angles with mean squared loss, \cite{Mahendran:ICCVW17} uses axis-angle or quaternion representations of 3D rotations and minimizes a geodesic loss during training. Our work is closest to \cite{Mahendran:ICCVW17} in that we also use an axis-angle representation and geodesic loss function while solving a 3D pose regression problem. However, there are three key differences between our work and \cite{Mahendran:ICCVW17}: (i) we solve the harder task of joint category and pose estimation, (ii) we design a new integrated architecture for the joint task, and (iii) our feature network architecture is based on residual networks \cite{He:CVPR16,He:arxiv16} whereas the feature network of \cite{Mahendran:ICCVW17} uses VGG-M \cite{Chatfield:BMVC14}. We refer the reader to \cite{Li:ICRB16} for a more detailed review of the current literature on 3D pose estimation using deep networks. 

\myparagraph{Joint object category and pose estimation} Braun \etal \cite{Braun:ITS16}, Massa \etal \cite{Massa:arxiv14,Massa:bmvc16} and O\~{n}oro-Rubio \etal \cite{Onoro:arxiv18} work on the problem of joint object detection and pose estimation, while Elhoseiny \etal \cite{Elhoseiny:ICML16} and Afifi \etal \cite{Afifi:VISAPP18} work on the problem of joint object category and pose estimation. However in all these works, the pose is restricted to just the azimuth or yaw angle. We, on the other hand, recover the full 3D rotation matrix which is a much harder problem. Also, these works all consider pose to be independent of category, \ie they set up a multi-task network that has separate branches for pose and category label, which are then treated independently. While \cite{Onoro:arxiv18} proposes various network architectures,  all of them are variations of shared networks (Fig.~\ref{fig:category_pose_networks}b) or independent networks (Fig.~\ref{fig:category_pose_networks}a). We, on the other hand, design an integrated network architecture (Fig.~2d) with a category-dependent pose network. 

\section{Joint object category and pose estimation}
\label{sec:model}
\vspace{-2mm}
In this section, we describe the proposed integrated architecture for joint object category and 3D pose estimation, which consists of a feature network, a category network and a category dependent pose network, as shown in Fig.~\ref{fig:category_pose_networks}d. In addition, we describe our loss functions and proposed training algorithm.

\vspace{-3mm}
\subsection{Integrated network architecture}
\label{sec:network}
\vspace{-2mm}

\myparagraph{Feature and category networks} Observe that the combination of our feature network and our categorization network resembles a standard categorization network in the image classification literature. Therefore, when designing our feature and categorization networks, we consider existing categorization architectures as our starting point. The recently introduced residual networks \cite{He:CVPR16,He:arxiv16} have been shown to work remarkably well for image classification as well as object detection. Taking inspiration from their success and good optimization properties \cite{Li:arxiv18}, we use residual networks (specifically the ResNet-50 network) in our work.

In image classification works, the last linear layer is considered a `categorization' network and everything before that is considered the `feature' network. In our case though, we are also interested in the 3D pose and such a splitting is not feasible because the image representations learned by such a feature network is highly specialized for categorization. We need to look at intermediate layers to find a suitable splitting point for our feature network and categorization network such that the image representations contain some pose information. Our experiments (in \S\ref{sec:resnet_analysis}) show that stage-4 is an appropriate splitting point and hence, we choose the ResNet-50 network upto stage-4 as our feature network and the stage-5 block of the ResNet-50 network as our categorization network.

\begin{figure}
	\vspace{-5mm}
	\centering
	\begin{subfigure}{0.55\linewidth}
		\includegraphics[width=\linewidth]{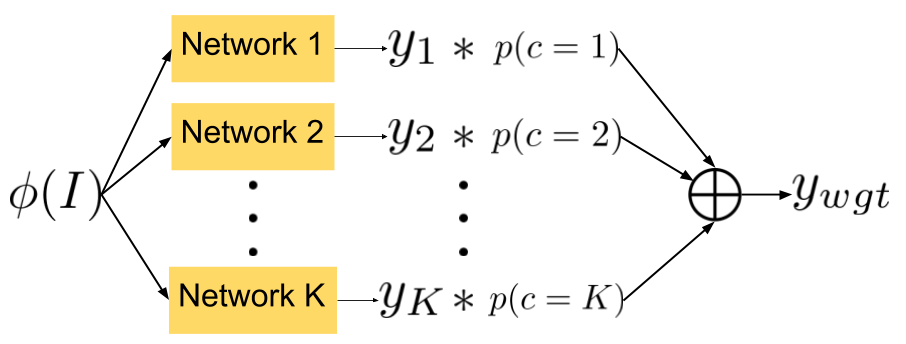}
		\caption{Overview with weighted fusion}
		\label{fig:category_dependent_pose_network}
	\end{subfigure}
	~
	\begin{subfigure}{0.4\linewidth}
		\scriptsize
		\centering
		{\color{red}\rotatebox{90}{\framebox[2cm][c]{$\phi(I)$}}}
		{\color{blue}\rotatebox{90}{\framebox[2cm][c]{FC1}}}
		{\color{magenta}\rotatebox{90}{\framebox[2cm][c]{BN1}}}
		{\color{cyan}\rotatebox{90}{\framebox[2cm][c]{ReLU}}}
		{\color{blue}\rotatebox{90}{\framebox[2cm][c]{FC2}}}
		{\color{magenta}\rotatebox{90}{\framebox[2cm][c]{BN2}}}
		{\color{cyan}\rotatebox{90}{\framebox[2cm][c]{ReLU}}}
		{\color{blue}\rotatebox{90}{\framebox[2cm][c]{FC3}}}
		{\color{cyan}\rotatebox{90}{\framebox[2cm][c]{$\pi \tanh$}}}
		\caption{3-layer FC pose network per object category (adapted from \cite{Mahendran:ICCVW17})}
		\label{fig:pose_network}
		\vspace{-5mm}
	\end{subfigure}
	
	\vspace{-2mm}
	\caption{Category dependent pose network}
	\vspace{-2em}
\end{figure}

\myparagraph{Category dependent pose network}
The proposed category dependent pose network is a collection of 3-layer fully connected (FC) pose networks (one per object category) that take in as input the output of the feature network and the probability vector output by the category network. An example is shown in Fig.~\ref{fig:category_dependent_pose_network}, where the output of the feature network $\phi(I)$ for image $I$ is input to $K$ FC pose networks. Each individual pose network (shown in Fig.~\ref{fig:pose_network}) consists of 3 FC layers (FC1, FC2 and FC3) with batch normalization (BN) and rectified linear units (ReLU) interspersed between them and a nonlinearity at the output. The outputs of these pose networks $\{\textbf{y}_i\}_{i=1}^K$ correspond to some representation of the object pose (see below for details). The category probability vector is denoted by $\textbf{p}$ where $\textbf{p}_i = P(c=i \mid I )$ is the probability of assigning the $i$-th category label to the image $I$. The $K$ category-dependent pose representations are fused together with the category probability vector $\textbf{p}$ to predict the final 3D pose using either a weighted or top-1 fusion strategy (see below for details). 

\myparagraph{Representation of 3D pose} Each pose network predicts a 3D pose (a rotation matrix R) for a given object category label c. We use the axis-angle representation, $R = \operatorname{expm}(\theta [\textbf{v}]_\times)$, where $\textbf{v}$ corresponds to the axis of rotation and $\theta$ is the angle of rotation. $[\textbf{v}]_\times$ is the skew-symmetric matrix generated from vector $\textbf{v}= [v_1, v_2, v_3]^T$, \ie $[\textbf{v}]_\times = \begin{pmatrix} 0 & -v_3 & v_2 \\ v_3 & 0 & -v_1 \\ -v_2 & v_1 & 0 \end{pmatrix}$. By restricting $\theta \in [0, \pi)$, we create a one-to-one correspondence between a rotation $R$ and its axis-angle vector $\textbf{y} = \theta \textbf{v}$. 
Let $\textbf{y}_i$ be the output of the $i$-th pose network. When the object category is known, we can choose the output $\textbf{y}_{c^*}$ corresponding to the true category $c^*$ and apply the exponential map $R_{c^*} = \operatorname{expm}([\textbf{y}_{c^*}]_\times)$ to obtain a rotation. Since in our case we do not know the category, the outputs of the $K$ pose networks need to be fused. Because the space of rotations is non-Euclidean, fusion is more easily done in the axis-angle space. Therefore, to obtain the final rotation, we first fuse the outputs of the $K$ pose networks and then apply the exponential~map.

\myparagraph{Pose fusion} 
Given $K$ category-dependent pose predictions $\{\mathbf{y}_i\}_{i=1}^K$ obtained by the $K$ pose networks and a category probability vector $\textbf{p}$ obtained from the category network, we propose two simple fusion strategies.

\mysubparagraph{Weighted fusion}
The fused pose output is a weighted sum of the individual pose predictions, with the weights determined by the category probability vector:
\vspace{-2mm}
\begin{equation}
\textbf{y}_{wgt} = \sum_i \textbf{y}_i \textbf{p}_i.
\label{eqn:weighted_pose}
\vspace{-2mm}
\end{equation}
This kind of fusion of pose outputs has some interesting properties:
\begin{enumerate}
	\item In the special case of know object category label, the category probability vector $\textbf{p} = \delta(c^*)$ and Eqn.~\ref{eqn:weighted_pose} naturally simplifies to $\textbf{y}_{wgt} = \textbf{y}_{c^*}$
	\item It is valid to define $\textbf{y}_{wgt} = \sum_i \textbf{y}_i \textbf{p}_i$ because a weighted sum of axis-angle vectors is still an axis-angle vector (axis-angle vectors are elements of the convex set $\{x \in \mathbb{R}^3 | \|x\|_2 < \pi \}$ and the weighted sum is a convex combination of elements of this convex set). On the other hand, a weighted sum of rotation matrices is not guaranteed to be a rotation matrix. 
\end{enumerate}

\vspace{-2mm}
\mysubparagraph{Top1 fusion} Instead of marginalizing out the category label, we can choose the final pose as the output corresponding to the mode of the category probability distribution. Effectively in this fusion, we are first estimating the object category label $\hat{c} = \operatorname{argmax}_i \textbf{p}_i$ and predicting pose accordingly 
\begin{equation}
\textbf{y}_{top1} = \textbf{y}_{\hat{c}}.
\label{eqn:top1_pose}
\vspace{-5mm}
\end{equation}

\vspace{-2mm}
\subsection{Loss functions}
\label{sec:loss}
In general, a loss function between ground-truth pose $R^*$, ground-truth category label $c^*$ and our network output $(R, c)$, can be expressed as $\mathcal{L}(R, c, R^*, c^*)$. A simple choice of the overall loss is to define it as a sum of a pose loss and a category loss, \ie $\mathcal{L}(R, c, R^*, c^*) = \mathcal{L}_{pose}(R(c), R^*) + \lambda \mathcal{L}_{category}(c, c^*)$, where the notation $R(c)$ explicitly encodes the fact that our pose output depends on the estimated category. We use the categorical cross-entropy loss as our category loss, and the geodesic distance between two rotation matrices $R_1$ and $R_2$, 
\vspace{-2mm}
\begin{equation}
\mathcal{L}_{pose}(R_1, R_2) = \frac{\|\log (R_1 R_2^T) \|_F}{\sqrt{2}},
\label{eqn:geodesic_loss}
\vspace{-2mm}
\end{equation}
as our pose loss. The pose loss between two axis-angle vectors $\textbf{y}_1$ and $\textbf{y}_2$ is now defined as $\mathcal{L}_p(\textbf{y}_1, \textbf{y}_2) \equiv \mathcal{L}(R_1, R_2)$ where $R_1$ and $R_2$ are the corresponding rotation matrices. Pose loss for the weighted fusion and top1 fusion pose outputs is now given by $\mathcal{L}_p(y^*, y_{wgt})$ and $\mathcal{L}_p(y^*, y_{top1})$ respectively.

In this paper, for our choice of representation and loss function, we do not observe a significant difference between the weighted and top1 fusion of pose outputs and report performance on both for all our experiments. However, for a different choice of representation, one might be better than the other. 

\vspace{-2mm}
\subsection{Network training}
\label{sec:network_training}
The overall network architecture has three sub-networks: the feature network (FN), the category network (CN) and the category dependent pose network (PN) consisting of one FC pose network per object category. A natural way to train this integrated network is to fix the feature network using weights from a pre-trained ResNet-50, train the category network \& each pose network independently and then finetune the overall network using our joint loss. This is called the ``balanced'' training approach where category and pose are treated in a balanced way. The fundamental problem with the balanced approach is that the feature network initialized with pre-trained weights is biased to categorization. And since we are solving a joint task with competing objectives where initialization is important, we end up with good categorization performance at the cost of pose estimation performance. Therefore, we propose a different approach to training the overall network called ``pose-first'' training. 

\begin{wrapfigure}{r}{0.5\linewidth}
	\centering
	\includegraphics[width=\linewidth]{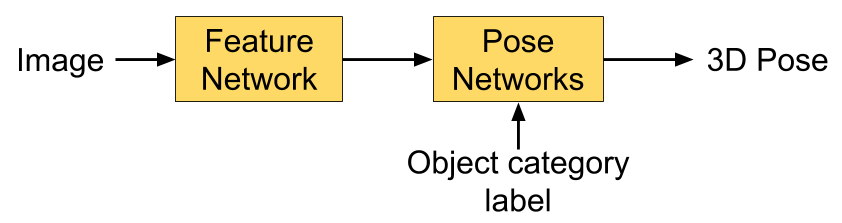}
	\vspace{-8mm}
	\caption{Training feature network + pose networks with oracle category network}
	\label{fig:pose_given_category}
	\vspace{-1cm}
\end{wrapfigure}
In this approach, we try to bias the feature network towards 3D pose estimation. We do this by first training our proposed network with an oracle category network \ie train the feature network and category dependent pose network, by minimizing the pose loss, with ground-truth object category labels (as shown in Fig.~\ref{fig:pose_given_category}). For every training image, the oracle category network returns a probability vector $\textbf{p} = \delta(c^*)$ that selects the pose network corresponding to its category label and updates only that pose network. In the process, we are also updating the feature network for the task of 3D pose estimation. With this updated feature network and category dependent pose network fixed, we now train the category network by minimizing the category loss. Then the overall network is finetuned end-to-end with both category and pose losses jointly. We also manually balance our training batch so that every batch has roughly the same number of training images per object category, which is an alternative implementation of the recommendation in \cite{Kokkinos:CVPR17} where they recommend asynchronous gradient updates to balance the variability of data across various
tasks and datasets. We show in our experiments in \S\ref{sec:joint_pose_and_category} that ``pose-first'' training achieves significantly better results for pose estimation compared to ``balanced'' training while achieving the marginally better results for category estimation. The ``pose-first'' training encodes what we set out to do, learn a 3D pose estimation system with known object category labels and then relax that constraint by replacing the oracle with a category estimation network.

\vspace{-1cm}
\begin{table}
	\caption{Table outlining the steps to train the overall network in two ways}
	\label{table:network_training}
	\centering
	\begin{tabular}{|c|c|c|}
		\hline
		Step & Balanced & Pose-first \\
		\hline
		1 & \multicolumn{2}{|c|}{Fix FN using weights from the pre-trained ResNet-50} \\
		2 & \multicolumn{2}{|c|}{Learn PN per object category independent of each other} \\
		\hline
		3 & Learn CN & Finetune FN+PN jointly with oracle CN \\
		4 & Finetune FN+CN+PN jointly & Learn CN with updated FN fixed \\
		5 & - & Finetune FN+CN+PN jointly \\
		\hline
	\end{tabular}
	\vspace{-8mm}
\end{table}

As mentioned earlier, we use the categorical cross-entropy loss for our category loss and the geodesic distance between rotation matrices for our pose loss. For evaluation, we use two metrics, (i) \textit{cat-acc}: the average accuracy in estimating the object category label (higher is better) and (ii) \textit{pose-err}: the median viewpoint error in degrees between the ground-truth and estimated rotations (lower is better). The viewpoint error between rotation matrices, Eqn.~\ref{eqn:geodesic_loss}, can be simplified using the Rodrigues' rotation formula to get viewpoint angle error (in degrees) between ground-truth rotation $R^*$ and predicted rotation $R$,
\vspace{-0.5em}
\begin{equation}
\Delta(R, R^*) = |\cos^{-1} \left[ \frac{ \operatorname{trace}(R^T R^*)-1}{2}\right]|.
\vspace{-0.5em}
\end{equation}
We use Adam optimizer \cite{Kingma:ICLR2014} in all our experiments and our code was written in Keras \cite{chollet2015keras} with TensorFlow backend \cite{tensorflow2015-whitepaper}.

\section{Results and Discussion}
\label{sec:experiments}

We first present the dataset we use for our experimental evaluation in \S\ref{sec:datasets}. Then, in \S\ref{sec:resnet_analysis}, we produce an investigation of features learned by a pre-trained ResNet-50 network for the task of 3D pose estimation. In \S \ref{sec:category_dependent_pose} we empirically verify a key assumption we make, that category dependent networks work better than category independent networks. Then, in \S\ref{sec:joint_pose_and_category} we report our experiments on the joint object category and pose estimation task. We demonstrate significant improvement upon state-of-the-art on the joint task and achieve performance comparable to methods that solve for 3D pose with known object category. We also present an extensive ablative analysis of many decisions choices like network architecture, feature representations, training protocol and relative weighting parameter. Finally, in \S\ref{sec:tsne} we present t-SNE visualizations of the features learned by our networks for the joint task and validate that the learned representations are suitable for both category and pose estimation tasks.

\subsection{Datasets}
\label{sec:datasets}
For our experiments, we use the challenging Pascal3D+ dataset (release version 1.1) \cite{Xiang:WACV14} which consists of images of 12 common rigid object categories of interest: aeroplane (aero), bicycle (bike), boat, bottle, bus, car, chair, diningtable (dtable), motorbike (mbike), sofa, train and tvmonitor (tv). The dataset includes Pascal VOC 2012 images \cite{PASCAL} and ImageNet images \cite{ImageNet} annotated with 3D pose annotations that describe the position of the camera with respect to the object in terms of azimuth, elevation, camera-tilt, distance, image-translation and focal-length. We use the ImageNet-training+validation images as our training data, Pascal-training images as our validation data and Pascal-validation images as our testing data. Like we mentioned earlier, we concentrate on the problem of joint object category and 3D pose estimation assuming we have bounding boxes around objects returned by an oracle. We use images that contain un-occluded and un-truncated objects that have been annotated with ground-truth bounding boxes. There are a total of 20,843 images that satisfy this criteria across these 12 categories of interest, with the number of images across the train-val-test splits detailed in Table \ref{table:num_images}. We use the 3D pose jittering data augmentation proposed in \cite{Mahendran:ICCVW17} and the rendered images \footnote{\url{https://shapenet.cs.stanford.edu/media/syn_images_cropped_bkg_overlaid.tar}}  provided in \cite{Su:ICCV15} to augment our training data.

\vspace{-2mm}
\begin{table}
	\caption{Number of images in Pascal3D+ v1.1 \cite{Xiang:WACV14} across various splits as well as rendered images provided by Su \etal \cite{Su:ICCV15}.
	}
	\label{table:num_images}
	\centering
	\begin{tabular}{|c|cccccccccccc|c|}
		\hline
		Category & aero & bike & boat & bottle & bus & car & chair & dtable & mbike & sofa & train & tv & Total \\
		\hline
		Train & 1765 & 794 & 1979 & 1303 & 1024 & 5287 & 967 & 737 & 634 & 601 & 1016 & 1195 & 17302 \\
		Val & 242 & 108 & 177 & 201 & 149 & 294 & 161 & 26 & 119 & 38 & 100 & 167 & 1782 \\
		Test & 244 & 112 & 163 & 177 & 144 & 262 & 180 & 17 & 127 & 37 & 105 & 191 & 1759 \\
		\hline
		Rendered & 198k & 200k & 199k & 200k & 199k & 195k & 197k & 196k & 200k & 200k & 200k & 199k & 2.381m \\
		\hline
	\end{tabular}
	\vspace{-5mm}
\end{table}

\subsection{Residual networks for 3D pose estimation}
\label{sec:resnet_analysis}
One of our contributions in this work is the use of the very popular residual networks designed by He \etal \cite{He:CVPR16,He:arxiv16} for the task of 3D pose estimation. As described in \S\ref{sec:network_training}, we initialize our overall network with pre-trained ResNet-50 networks. Before we present our experimental results though, we would like to answer the question: Can we adapt residual networks trained for the task of image classification to the task of pose estimation? Networks learned for the task of image classification are trained to be invariant to pose. However, previous works like \cite{Elhoseiny:ICML16} (for AlexNet \cite{Krizhevsky:NIPS12}) and \cite{Mahendran:ICCVW17} (for VGG-M \cite{Chatfield:BMVC14}) have shown that features extracted from intermediate layers retain information relevant to pose estimation. For our specific case, we are also interested in knowing at what intermediate stage does a pre-trained ResNet-50 have relevant pose information.

\begin{wrapfigure}{r}{0.52\linewidth}
	\vspace{-8mm}
	\centering
	\includegraphics[width=\linewidth]{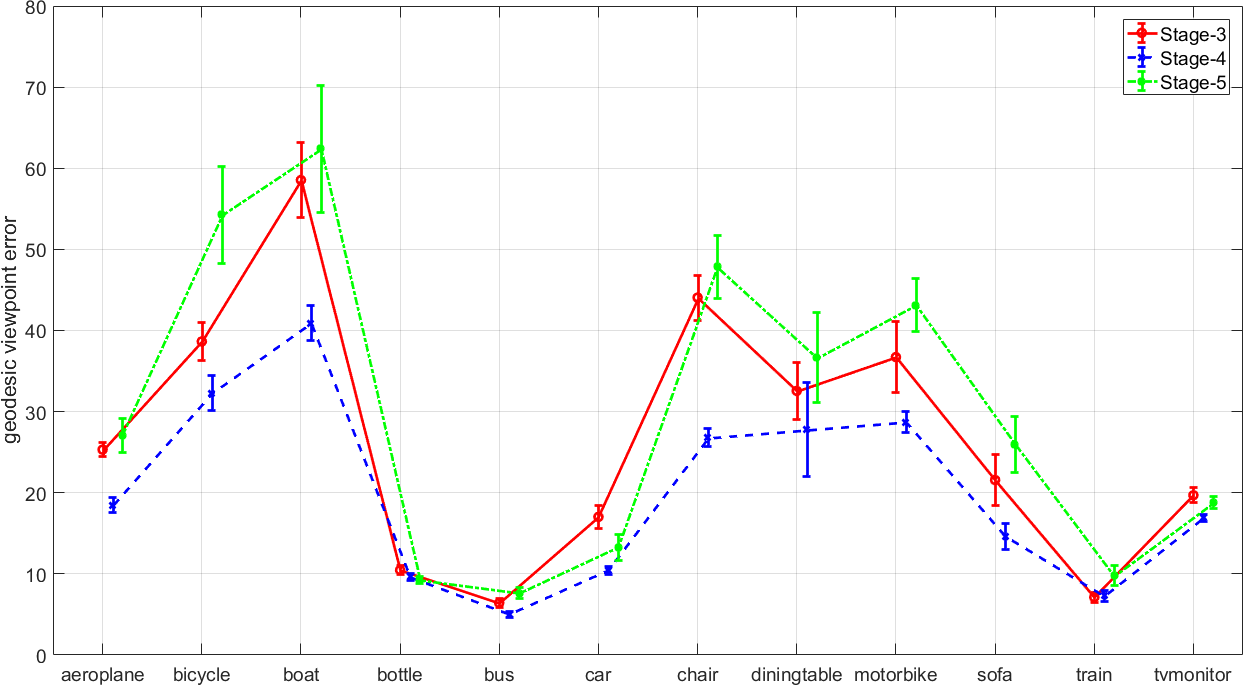}
	\caption{Median viewpoint error (in degrees) with features extracted from different stage of the pre-trained ResNet-50. Lower is better. Best seen in color.}
	\label{fig:resnet50_features}
	\vspace{-5mm}
\end{wrapfigure}
We extract features at different stages of the ResNet-50 architecture with pre-trained weights. We then learn 3-layer FC networks per object category of size $d$-1000-500-3 using these features. As can be seen in Fig.~\ref{fig:resnet50_features} and Table~\ref{table:resnet_features}, we find that features extracted at the end of stage-4 are better than the features extracted at the end of stages-3 and 5. This is consistent with previous findings that show that (i) features become more specialized for the task they were trained for (image classification in this case) the deeper they are in the network, which explains why stage-4 features are better than stage-5, and (ii) deeper layers capture more complex information compared to simple edge detectors at the first few layers, which explains why stage-4 features are better than stage-3.

\begin{table}
	\vspace{-6mm}
	\centering
	\setlength{\tabcolsep}{0.7mm}
	\scriptsize
	\caption{Median viewpoint error (in degrees) after learning pose networks using features extracted from pre-trained networks. Pose networks are of size 512/1024/2048-1000-500-3 for ResNet-50 Stages-3/4/5 respectively. Lower is better and best results in bold.}
	\label{table:resnet_features}
	\begin{tabular}{|cc|cccccccccccc|c|}
		\hline
		Network & Feats. & aero & bike & boat & bottle & bus & car & chair & dtable & mbike & sofa & train & tv & mean \\
		\hline
		\multirow{3}{*}{ResNet-50} & Stage-3 & 25.3 & 38.6 & 58.6 & 10.5 & 6.4 & 17.0 & 44.1 & 32.6 & 36.7 & 21.6 & \textbf{7.1} & 19.7 & 26.51 \\ 
		& Stage-4 & \textbf{18.4} & \textbf{32.3} & \textbf{40.9} & 9.6 & \textbf{5.0} & \textbf{10.4} & \textbf{26.8} & \textbf{27.8} & \textbf{28.7} & \textbf{14.6} & 7.3 & \textbf{16.9} & \textbf{19.91} \\ 
		& Stage-5 & 27.1 & 54.3 & 62.4 & \textbf{9.2} & 7.6 & 13.3 & 47.8 & 36.7 & 43.1 & 26.0 & 9.8 & 18.8 & 29.67 \\ 
		\hline
	\end{tabular}
	\vspace{-8mm}
\end{table}

\subsection{Category-dependent pose networks}
\label{sec:category_dependent_pose}
An implicit assumption we have made in our work is that our choice of category-dependent pose network architecture (with per-category pose networks) is better than the choice of a category-independent pose network (the choice of \cite{Elhoseiny:ICML16,Braun:ITS16}). 
In our architecture, the feature network is shared between all object categories and the pose networks are specific to each category. \cite{Elhoseiny:ICML16} discusses where the branching between category and pose estimation tasks should occur in their early branching (EBM) and late branching (LBM) models, but they do not discuss why they choose a category-independent pose network. We now validate our decision choice of category-dependent pose networks empirically. For this experiment, we use the features extracted from ResNet-50 stage-4. We then learn twelve 3-layer pose networks, one for each object category, of size 1024-1000-500-3 (Row 1 of Table~\ref{table:category_dependent_pose}). To compare with a category independent pose network, we use these same features to learn a single pose network of size 1024-12000-500-3 (Row 2 of Table~\ref{table:category_dependent_pose}). Note that the intermediate layer is of size 12000 to have roughly the same number of total parameters as the 12 independent pose networks. We also show performance on two smaller intermediate sizes and as can be seen in Table~\ref{table:category_dependent_pose}, solving for the pose in a per-category manner is better. 

\begin{table}
	\vspace{-6mm}
	\caption{Median viewpoint error (in degrees) for category-dependent and category-independent pose networks. For the category-dependent network, there are 12 networks of size 1024-1000-500-3. We evaluate three sizes of category-independent networks: 1024-12000-500-3 (12k), 1024-5000-500-3 (5k) and 1024-1000-500-3 (1k). Lower is better and best results in bold.}
	\label{table:category_dependent_pose}
	\centering
	\setlength{\tabcolsep}{0.7mm}
	\scriptsize
	\begin{tabular}{|c|cccccccccccc|c|}
		\hline
		Type & aero & bike & boat & bottle & bus & car & chair & dtable & mbike & sofa & train & tv & Mean \\
		\hline
		Cat-dep. & \textbf{18.4} & \textbf{32.3} & \textbf{40.9} & \textbf{9.6} & \textbf{5.0} & \textbf{10.4} & \textbf{26.8} & 27.8 & \textbf{28.7} & 14.6 & \textbf{7.3} & \textbf{16.9} & \textbf{19.91} \\ 
		\hline
		Cat-ind.(12k) & 23.9 & 40.9 & 54.7 & 12.1 & 5.8 & 11.6 & 41.8 & 26.9 & 31.0 & 20.3 & 10.0 & 18.9 & 24.83 \\ 
		Cat-ind.(5k) & 22.3 & 39.6 & 52.6 & 12.4 & 5.5 & 12.2 & 44.1 & 27.4 & 32.4 & \textbf{14.6} & 10.2 & 18.7 & 24.3 \\ 
		Cat-ind.(1k) & 22.9 & 36.6 & 55.4 & 10.8 & 5.5 & 10.6 & 36.9 & \textbf{26.3} & 32.4 & 18.5 & 9.4 & 18.8 & 23.69 \\
		\hline 
	\end{tabular}
	\vspace{-8mm}
\end{table}

\subsection{Joint object category and pose estimation}
\label{sec:joint_pose_and_category}
We now present the results of our experiments on the task of joint object category and pose estimation. As mentioned earlier in Sec.\ref{sec:network_training} we train the overall network using the ``pose-first'' approach with both weighted and top1 fusion strategies of Sec.~\ref{sec:network}. To evaluate our performance we report the object category estimation accuracy (cat-acc) and the median viewpoint error (pose-err), across all object categories. As can be seen in Table~\ref{table:joint_results}, we achieve close to $90\%$ category estimation accuracy and slightly more than $16^\circ$ median viewpoint error averaged across object categories in both models when training on only real images. Using rendered images to augment training data leads to significant improvement in pose estimation performance, with pose viewpoint error decreasing by $\sim2.4^\circ$ in the Weighted model and $\sim2.7^\circ$ in the Top1 model, and an improvement of $\sim2$\% in the category estimation accuracy for both models. Note that rendered images are valid only for the pose estimation part of our problem and not the category estimation part. In the ``pose-first'' training method, we use these rendered images in steps 2\&3 of Table~\ref{table:network_training}. This is used to initialize the joint network which is subsequently trained using only real (original + flipped) images from PASCAL3D+. Our results are consistent with those of \cite{Su:ICCV15} and \cite{Mahendran:ICCVW17} who also observed improved performance by using rendered images for data augmentation.

\begin{table}
	\vspace{-2mm}
	\tiny
	\setlength{\tabcolsep}{0.7mm}
	\centering
	\caption{Object category estimation accuracy (percentage, higher is better) and pose viewpoint error (degress, lower is better) for experiments with joint networks using real and rendered images. Best results are in bold.}
	\label{table:joint_results}
	\begin{tabular}{|c|c|c|cccccccccccc|c|}
		\hline
		Metric & Data & Model & aero & bike & boat & bottle & bus & car & chair & dtable & mbike & sofa & train & tv & mean \\
		\hline
		\multirow{4}{*}{pose-err} & \multirow{2}{*}{\makecell{Only real \\ images}} & Weighted & 13.6 & 21.3 & 34.8 & 9.0 & 3.4 & 7.8 & 26.6 & \textbf{20.8} & 17.6 & 15.8 & 6.9 & 15.2 & 16.07 \\ 
		& & Top1 & 13.4 & 22.2 & 33.5 & 9.2 & 3.3 & 7.7 & 26.2 & 24.0 & 17.8 & 16.5 & \textbf{6.6} & 15.2 & 16.29 \\ 
		\cline{2-16}
		& \multirow{2}{*}{\makecell{Real and \\ rendered}} & Weighted & 11.5 & \textbf{15.7} & 30.5 & 9.0 & 2.9 & 6.8 & 16.2 & 22.5 & 14.4 & \textbf{13.8} & 7.3 & \textbf{13.4} & 13.67 \\ 
		& & Top1 & \textbf{10.2} & 17.1 & \textbf{29.2} & \textbf{8.1} & \textbf{2.6} & \textbf{6.0} & \textbf{13.8} & 26.7 & \textbf{14.1} & 14.3 & 7.1 & 13.5 & \textbf{13.56} \\ 
		\hline
		\hline
		\multirow{4}{*}{cat-acc} & \multirow{2}{*}{\makecell{Only real \\ images}} & Weighted & 0.94 & 0.89 & 0.95 & 0.98 & 0.96 & 0.94 & 0.83 & 0.67 & 0.93 & 0.78 & 0.94 & 0.93 & 0.8944 \\ 
		& & Top1 & \textbf{0.97} & 0.87 & 0.94 & 0.96 & 0.96 & 0.95 & 0.84 & 0.62 & \textbf{0.96} & 0.78 & 0.94 & 0.93 & 0.8930 \\ 
		\cline{2-16}
		& \multirow{2}{*}{\makecell{Real and \\ rendered}} & Weighted & 0.96 & \textbf{0.94} & 0.97 & \textbf{1.00} & 0.95 & \textbf{0.96} & 0.87 & 0.71 & \textbf{0.96} & 0.77 & \textbf{0.95} & \textbf{0.95} & 0.9150 \\ 
		& & Top1 & 0.96 & 0.92 & \textbf{0.98} & 0.98 & \textbf{0.97} & \textbf{0.96} & \textbf{0.89} & \textbf{0.76} & 0.93 & \textbf{0.82} & 0.88 & \textbf{0.95} & \textbf{0.9181} \\ 
		\hline
	\end{tabular}
	\vspace{-5mm}
\end{table}

\myparagraph{Comparison with State-of-the-art} To the best of our knowledge, Elhoseiny \etal \cite{Elhoseiny:ICML16} are the current state-of-the-art on the PASCAL3D+ dataset for the task of joint object category and azimuth estimation given ground-truth bounding boxes. We do not use any rendered images in these experiments to ensure a fair comparison. They report the performance of their models using the following metrics: (i) P\%($<22.5^\circ/45^\circ$): percentage of images that have pose error less than $22.5^\circ/45^\circ$ and (ii) AAAI: $1 - [\min(|err|, 2\pi - |err|)/\pi ]$. We evaluate our models using these metrics for both azimuth error $|az - az^*|$ and 3D pose error $\Delta(R, R^*)$. For azimuth estimation, we predict 3D rotation matrix and then retrieve the azimuth angle. For a fair comparison between their method and ours, we re-implemented their algorithm with our network architecture \ie feature network is ResNet-50 upto Stage-4, category network is ResNet-50 Stage-5 and pose network is a 3-layer FC network of size 1024-1000-500-3. This size of the pose network was chosen based on Table~\ref{table:category_dependent_pose}. As can be seen in Tables~\ref{table:joint_results_comp} and \ref{table:elhoseiny_comp}, we perform significantly better than \cite{Elhoseiny:ICML16} in pose estimation accuracy under different metrics while performing marginally worse ($0.3-0.6$\%) in category accuracy.

\begin{table}
	\vspace{-8mm}
	\tiny
	\caption{Comparing our joint networks with weighted (wgt) and top1 pose output with \cite{Elhoseiny:ICML16}* (our re-implementation of \cite{Elhoseiny:ICML16}, see text for details). Higher is better for the cat-acc, lower is better for pose-err metric. Best results in bold.}
	\label{table:joint_results_comp}
	\centering
	\setlength{\tabcolsep}{1mm}
	\begin{tabular}{|c|c|cccccccccccc|c|}
		\hline
		Metric & Model & aero & bike & boat & bottle & bus & car & chair & dtable & mbike & sofa & train & tv & Mean \\
		\hline
		\multirow{3}{*}{pose-err} & \cite{Elhoseiny:ICML16}* & 17.7 & 24.7 & 41.6 & 9.9 & 3.6 & 12.2 & 31.9 & 20.8 & 20.2 & 26.7 & 6.9 & 16.0 & 19.35 \\ 
		& Ours-Wgt & 11.5 & \textbf{15.7} & 30.5 & 9.0 & 2.9 & 6.8 & 16.2 & \textbf{22.5} & 14.4 & \textbf{13.8} & 7.3 & \textbf{13.4} & 13.67 \\ 
		& Ours-Top1 & \textbf{10.2} & 17.1 & \textbf{29.2} & \textbf{8.1} & \textbf{2.6} & \textbf{6.0} & \textbf{13.8} & 26.7 & \textbf{14.1} & 14.3 & \textbf{7.1} & 13.5 & \textbf{13.56} \\ 
		\hline
		\hline
		\multirow{3}{*}{cat-acc} & \cite{Elhoseiny:ICML16}* & 0.95 & \textbf{0.94} & \textbf{0.98} & 0.98 & \textbf{0.99} & \textbf{0.96} & \textbf{0.89} & 0.57 & \textbf{0.96} & \textbf{0.90} & \textbf{0.97} & \textbf{0.96} & \textbf{0.9215} \\ 
		& Ours-Wgt & \textbf{0.96} & \textbf{0.94} & 0.97 & \textbf{1.00} & 0.95 & \textbf{0.96} & 0.87 & 0.71 & \textbf{0.96} & 0.77 & 0.95 & 0.95 & 0.9150 \\ 
		& Ours-Top1 & \textbf{0.96} & 0.92 & \textbf{0.98} & 0.98 & 0.97 & \textbf{0.96} & \textbf{0.89} & \textbf{0.76} & 0.93 & 0.82 & 0.88 & 0.95 & 0.9181 \\ 
		\hline
	\end{tabular}
	\vspace{-8mm}
\end{table}

\begin{table}
	\vspace{-6mm}
	\scriptsize
	\setlength{\tabcolsep}{1.5mm}
	\caption{Comparing our joint networks with weighted (wgt) and top1 pose output with \cite{Elhoseiny:ICML16}* under their metrics. Higher is better, Best results in bold.}
	\label{table:elhoseiny_comp}
	\centering
	\begin{tabular}{|c|c|cccc|}
		\hline
		Error & Model & cat-acc & P\%($<22.5^\circ$) & P\%($<45^\circ$) & AAAI \\
		\hline
		\multirow{3}{*}{$\Delta(R, R^*)$} & \cite{Elhoseiny:ICML16}* & \textbf{0.9215} & 0.5759 & 0.7647 & 0.8128 \\
		& Ours-Weighted & 0.9150 & 0.6853 & 0.8370 & 0.8495 \\
		& Ours-Top1 & 0.9181 & \textbf{0.7054} & \textbf{0.8436} & \textbf{0.8526} \\
		\hline
		\hline
		\multirow{4}{*}{$\Delta(az, az^*)$} & \cite{Elhoseiny:ICML16}* & \textbf{0.9215} & 0.5009 & 0.6939 & 0.7692 \\
		& Ours-Weighted & 0.9150 & 0.6627 & 0.7797 & 0.8130 \\
		& Ours-Top1 & 0.9181 & \textbf{0.7287} & \textbf{0.8170} & \textbf{0.8333} \\
		\cline{2-6}
		& \cite{Elhoseiny:ICML16} & 0.8379 & 0.5189 & 0.6074 & 0.7539 \\
		\hline
	\end{tabular}
	\vspace{-5mm}
\end{table}

In Table~\ref{table:pose_estimation}, we compare with state-of-the-art methods on pose estimation with known object category and observe that we achieve competitive performance even though we are solving a harder problem of joint object category and pose estimation. We would like to explicitly mention that, by solving the joint task, we are not trying to achieve improved performance relative to pose estimation with known object category. As we shall show later, the fine-tuning step where we jointly minimize pose and category losses shows a trade-off between pose estimation and category estimation performance, further proving our intuition that pose estimation and categorization are competing (not synergistic) tasks. Our motivation is to relax the known object category label constraint and still achieve comparable results.

\begin{table}
	\vspace{-8mm}
	\scriptsize
	\centering
	\setlength{\tabcolsep}{0.3mm}
	\caption{Median viewpoint error (in degrees, lower is better) across 12 object categories on Pascal3D+ dataset. First four rows are 3D pose estimation methods with known object category and the last two rows (Ours) are joint object category and 3D pose estimation methods. Best results are highlighted in bold and second best are in red. Best seen in color.}
	\label{table:pose_estimation}
	\begin{tabular}{|c|cccccccccccc|c|}
		\hline
		Methods & aero & bike & boat & bottle & bus & car & chair & dtable & mbike & sofa	& train & tv & Mean \\
		\hline
		Viepoints\&Keypoints \cite{Tulsiani:CVPR15} & 13.8 & 17.7 & \textbf{21.3} & 12.9 & 5.8 & 9.1 & 14.8 & 15.2 & 14.7 & 13.7 & 8.7 & 15.4 & 13.59 \\ 
		Render-for-CNN \cite{Su:ICCV15} & 15.4 & \textcolor{red}{14.8} & 25.6 & 9.3 & 3.6 & \textcolor{red}{6.0} & \textbf{9.7} & \textbf{10.8} & 16.7 & \textbf{9.5} & \textbf{6.1} & \textcolor{red}{12.6} & \textcolor{red}{11.67} \\ 
		3D-pose-regression \cite{Mahendran:ICCVW17} & 13.97 & 21.07 & 35.52 & 8.99 & 4.08 & 7.56 & 21.18 & 17.74 & 17.87 & \textcolor{red}{12.70} & 8.22 & 15.68 & 15.38 \\ 
		Multibin \cite{Mousavian:CVPR17} & 13.6 & \textbf{12.5} & \textcolor{red}{22.8} & \textcolor{red}{8.3} & 3.1 & \textbf{5.8} & \textcolor{red}{11.9} & \textcolor{red}{12.5} & \textbf{12.3} & 12.8 & \textcolor{red}{6.3} & \textbf{11.9} & \textbf{11.1} \\
		\hline
		\hline
		Ours-Weighted & \textcolor{red}{11.5} & 15.7 & 30.5 & 9.0 & \textcolor{red}{2.9} & 6.8 & 16.2 & 22.5 & 14.4 & 13.8 & 7.3 & 13.4 & 13.67 \\ 
		Ours-Top1 & \textbf{10.2} & 17.1 & 29.2 & \textbf{8.1} & \textbf{2.6} & \textcolor{red}{6.0} & 13.8 & 26.7 & \textcolor{red}{14.1} & 14.3 & 7.1 & 13.5 & 13.56 \\ 
		\hline
	\end{tabular}
	\vspace{-5mm}
\end{table}

\myparagraph{Choice of Network Architecture} In \S\ref{sec:introduction}, we introduced four possible network architectures to solve a joint task and advocated for Integrated network over other choices. We experimentally validate that choice by comparing it with a Sequential network and Shared network. Like we mention earlier, \cite{Elhoseiny:ICML16} is a Shared network. For a Sequential network, we use a second ResNet-50 network and finetune different parts of the network for categorization on the Pascal3D+ dataset.

\begin{table}
	\vspace{-8mm}
	\caption{Comparison between Sequential, Shared and Integrated networks.}
	\label{table:network_comp}
	\tiny
	\setlength{\tabcolsep}{0.7mm}
	\centering
	\begin{tabular}{|c|c|c|cccccccccccc|c|}
		\hline
		Metric & Network & Model & aero & bike & boat & bottle & bus & car & chair & dtable & mbike & sofa & train & tv & Mean \\
		\hline
		\multirow{6}{*}{pose-err} & \multirow{3}{*}{Sequential} & Everything & 11.7 & 17.5 & 27.4 & 8.4 & 3.3 & \textbf{5.9} & 14.5 & 24.8 & 15.0 & \textbf{11.1} & 6.3 & \textbf{13.2} & \textbf{13.25} \\ 
		& & FC & 11.7 & 17.7 & \textbf{27.1} & 8.5 & 3.2 & 6.4 & 16.7 & 24.8 & 15.0 & \textbf{11.1 }& 6.9 & \textbf{13.2} & 13.54 \\
		& & FC+Stage5 & 11.7 & 17.6 & 27.3 & 8.2 & 3.2 & \textbf{5.9} & 14.5 & 24.8 & 14.3 & 12.7 & \textbf{6.3} & \textbf{13.2} & 13.30 \\
		\cline{2-16}
		& \multirow{2}{*}{Integrated} & Ours-Wgt & 11.5 & \textbf{15.7} & 30.5 & 9.0 & 2.9 & 6.8 & 16.2 & 22.5 & 14.4 & 13.8 & 7.3 & 13.4 & 13.67 \\
		& & Ours-Top1 & \textbf{10.2} & 17.1 & 29.2 & \textbf{8.1} & \textbf{2.6} & 6.0 & \textbf{13.8} & 26.7 & \textbf{14.1} & 14.3 & 7.1 & 13.5 & 13.56 \\ 
		\cline{2-16}
		& Shared & \cite{Elhoseiny:ICML16}* & 17.7 & 24.7 & 41.6 & 9.9 & 3.6 & 12.2 & 31.9 & \textbf{20.8} & 20.2 & 26.7 & 6.9 & 16.0 & 19.35 \\
		\hline
		\hline
		\multirow{6}{*}{cat-acc} & \multirow{3}{*}{Sequential} & Everything & 0.96 & 0.92 & \textbf{0.98} & 0.98 & 0.95 & \textbf{0.97} & 0.88 & \textbf{0.76} & 0.93 & 0.85 & \textbf{0.97} & 0.97 & 0.9275 \\ 
		& & FC & 0.94 & 0.92 & 0.96 & 0.97 & 0.97 & 0.88 & 0.81 & 0.62 & 0.93 & \textbf{0.95} & 0.95 & 0.95 & 0.9033 \\ 
		& & FC+Stage5 & \textbf{0.98} & 0.92 & \textbf{0.98} & 0.99 & 0.97 & \textbf{0.97} & \textbf{0.90} & 0.67 & \textbf{0.96} & 0.92 & \textbf{0.99} & 0.96 & \textbf{0.9348} \\
		\cline{2-16}
		& \multirow{2}{*}{Integrated} & Ours-Wgt & 0.96 & \textbf{0.94} & 0.97 & \textbf{1.00} & 0.95 & 0.96 & 0.87 & 0.71 & \textbf{0.96} & 0.77 & 0.95 & 0.95 & 0.9150 \\
		& & Ours-Top1 & 0.96 & 0.92 & \textbf{0.98} & 0.98 & 0.97 & 0.96 & 0.89 & \textbf{0.76} & 0.93 & 0.82 & 0.88 & 0.95 & 0.9181 \\ 
		\cline{2-16}
		& Shared & \cite{Elhoseiny:ICML16}* & 0.95 & \textbf{0.94} & \textbf{0.98} & 0.98 & \textbf{0.99} & 0.96 & 0.89 & 0.57 & \textbf{0.96} & 0.90 & 0.97 & 0.96 & 0.9215 \\
		\hline
	\end{tabular}
	\vspace{-5mm}
\end{table}

As can be seen in Table~\ref{table:network_comp}, our Integrate networks are clearly better than the Shared network. They are comparable to Sequential networks in pose estimation performance and slightly worse in categorization accuracy while being significantly cheaper computationally. The Sequential network has to maintain two ResNet-50 networks compared to a single one for our Integrated networks.

\myparagraph{Choice of training protocol} In Sec.~\ref{sec:network_training}, we proposed two approaches to train the overall network. As can be seen in Table~\ref{table:joint_approach}, averaged across all object categories, the joint models trained using the ``pose-first'' approach are better than those trained using the ``balanced'' one. This means that learning the best possible feature + pose models for the task of 3D pose estimation first and then relaxing them to solve the joint category and pose estimation task is better than solving category and pose estimation tasks independently with fixed feature network and then training everything jointly. 

\begin{table}
	\vspace{-2mm}
	\centering
	\tiny
	\setlength{\tabcolsep}{0.7mm}
	\caption{Object category estimation accuracy (percentage, higher is better) and pose viewpoint error (degrees, lower is better) using the two training approaches described in Sec.~\ref{sec:network_training} for Real Images. Best results in bold.}
	\label{table:joint_approach}
	\begin{tabular}{|c|c|c|cccccccccccc|c|}
		\hline
		Metric & Expt. & Type & aero & bike & boat & bottle & bus & car & chair & dtable & mbike & sofa & train & tv & mean \\
		\hline
		\multirow{4}{*}{pose-err} & \multirow{2}{*}{Balance} & Weighted & 13.9 & 24.1 & 35.4 & 9.6 & 4.1 & 8.8 & 33.5 & 24.1 & 19.5 & 17.7 & 7.6 & 15.9 & 17.86 \\ 
		& & Top1 & 13.7 & 23.9 & \textbf{32.4} & 10.6 & 3.8 & 8.0 & 32.5 & 26.3 & 17.8 & 18.0 & 7.6 & 15.4 & 17.51 \\ 
		& \multirow{2}{*}{Pose-first} & Weighted & 13.6 & \textbf{21.3} & 34.8 & \textbf{9.0} & 3.4 & 7.8 & 26.6 & \textbf{20.8} & \textbf{17.6} & \textbf{15.8} & 6.9 & \textbf{15.2} & \textbf{16.07} \\ 
		& & Top1 & \textbf{13.4} & 22.2 & 33.5 & 9.2 & \textbf{3.3} & \textbf{7.7} & \textbf{26.2} & 24.0 & 17.8 & 16.5 & \textbf{6.6} & \textbf{15.2} & 16.29 \\ 
		\hline
		\hline
		\multirow{4}{*}{cat-acc} & \multirow{2}{*}{Balance} & Weighted & 0.92 & \textbf{0.89} & 0.96 & 0.96 & 0.95 & 0.81 & 0.73 & 0.71 & 0.94 & \textbf{0.87} & \textbf{0.96} & \textbf{0.96} & 0.8889 \\ 
		& & Top1 & 0.90 & \textbf{0.89} & \textbf{0.97} & 0.88 & 0.94 & 0.79 & 0.70 & \textbf{0.73} & 0.95 & \textbf{0.87} & 0.95 & 0.95 & 0.8760 \\ 
		& \multirow{2}{*}{Pose-first} & Weighted & 0.94 & \textbf{0.89} & 0.95 & \textbf{0.98} & \textbf{0.96} & 0.94 & 0.83 & 0.67 & 0.93 & 0.78 & 0.94 & 0.93 & \textbf{0.8944} \\ 
		& & Top1 & \textbf{0.97} & 0.87 & 0.94 & 0.96 & \textbf{0.96} & \textbf{0.95} & \textbf{0.84} & 0.62 & \textbf{0.96} & 0.78 & 0.94 & 0.93 & 0.8930 \\ 
		\hline
	\end{tabular}
	\vspace{-5mm}
\end{table}

\myparagraph{Effect of finetuning the overall network with the joint losses} The tasks of object category estimation (which requires the network to be invariant to pose) and 3D pose estimation (which is invariant to object sub-category) are competing with each other and during joint training, the feature network tries to learn a representation that is suitable for both tasks. In doing so, we expect a the trade-off between cat-acc and pose-err performance. This trade-off is observed explicitly when trying to learn the overall model using the ``balanced'' approach. We first do steps 1-3, where we fix the feature network to the pre-trained weights and train the pose and category networks independently. Rows 1-3 of Table~\ref{table:joint_finetuning} show that the original feature network with pre-trained weights has features that are very good for category estimation but not for pose estimation. We then do step 4, where we finetune the overall network with our joint loss. As can be seen in Table~\ref{table:joint_finetuning}, there is a trade-off between the two tasks of object category estimation \& 3D pose estimation and we lose some category estimation accuracy while improving the pose estimation performance. 

\begin{table}
	\vspace{-8mm}
	\tiny
	\centering
	\setlength{\tabcolsep}{0.7mm}
	\caption{Object category estimation accuracy (percentage, higher is better) and pose viewpoint error (degrees, lower is better) before (Steps 1-3) and after (Step 4) finetuning the overall network in ``balanced'' training. Best results in bold.}
	\label{table:joint_finetuning}
	\begin{tabular}{|c|c|c|cccccccccccc|c|}
		\hline
		Metric & Expt. & Type & aero & bike & boat & bottle & bus & car & chair & dtable & mbike & sofa & train & tv & mean \\
		\hline
		\multirow{4}{*}{pose-err} & \multirow{2}{*}{Before} & Weighted & 25.5 & 39.0 & 47.6 & 13.4 & 9.7 & 15.2 & 34.6 & 38.2 & 37.5 & 21.5 & 9.6 & 16.6 & 25.70 \\ 
		& & Top1 & 25.5 & 38.9 & 49.5 & 13.4 & 9.4 & 14.5 & 35.8 & 38.2 & 37.2 & 21.5 & 8.9 & 16.7 & 25.79 \\ 
		& \multirow{2}{*}{After} & Weighted & 13.9 & 24.1 & 35.4 & \textbf{9.6} & 4.1 & 8.8 & 33.5 & \textbf{24.1} & 19.5 & \textbf{17.7} & \textbf{7.6} & 15.9 & 17.86 \\ 
		& & Top1 & \textbf{13.7} & \textbf{23.9} & \textbf{32.4} & 10.6 & \textbf{3.8} & \textbf{8.0} & \textbf{32.5} & 26.3 & \textbf{17.8} & 18.0 & \textbf{7.6} & \textbf{15.4} & \textbf{17.51} \\ 
		\hline
		\hline
		\multirow{3}{*}{cat-acc} & Before & Both & \textbf{0.97} & \textbf{0.91} & \textbf{0.97} & \textbf{0.98} & \textbf{0.95} & \textbf{0.92} & \textbf{0.90} & 0.71 & \textbf{0.97} & 0.85 & 0.95 & \textbf{0.99} & \textbf{0.9226} \\ 
		& \multirow{2}{*}{After} & Weighted & 0.92 & 0.89 & 0.96 & 0.96 & \textbf{0.95} & 0.81 & 0.73 & 0.71 & 0.94 & \textbf{0.87} & \textbf{0.96} & 0.96 & 0.8889 \\ 
		& & Top1 & 0.90 & 0.89 & \textbf{0.97} & 0.88 & 0.94 & 0.79 & 0.70 & \textbf{0.73} & 0.95 & \textbf{0.87} & 0.95 & 0.95 & 0.8760 \\ 
		\hline
	\end{tabular}	
	\vspace{-5mm}
\end{table}

\myparagraph{Choice of $\lambda$} In Table~\ref{table:joint_lambda}, we compare the performance of models learned with different choices of $\lambda$. The $\lambda$ parameter controls the relative importance of the category loss (categorical cross-entropy) w.r.t. the pose loss (geodesic loss). Recall that $\mathcal{L}(R, c, R^*, c^*) = \mathcal{L}_{pose}(R(c), R^*) + \lambda \mathcal{L}_{category}(c, c^*)$. The smaller value of $\lambda=0.1$ lead to better joint models and unless mentioned otherwise all the models were trained with that choice of $\lambda$.

\begin{table}
	\tiny
	\centering
	\setlength{\tabcolsep}{0.7mm}
	\caption{Object category estimation accuracy (percentage, higher is better) and pose viewpoint error (degrees, lower is better) under the two joint losses for different $\lambda$ trained using the pose-first approach. Best results in bold.}
	\label{table:joint_lambda}
	\begin{tabular}{|c|c|c|cccccccccccc|c|}
		\hline
		Metric & Expt. & Type & aero & bike & boat & bottle & bus & car & chair & dtable & mbike & sofa & train & tv & mean \\
		\hline
		\multirow{4}{*}{pose-err} & \multirow{2}{*}{$\lambda=0.1$} & Weighted & 13.6 & \textbf{21.3} & 34.8 & \textbf{9.0} & 3.4 & 7.8 & 26.6 & 20.8 & \textbf{17.6} & \textbf{15.8} & 6.9 & \textbf{15.2} & \textbf{16.07} \\ 
		& & Top1 & 13.4 & 22.2 & \textbf{33.5} & 9.2 & \textbf{3.3} & \textbf{7.7} & \textbf{26.2} & 24.0 & 17.8 & 16.5 & \textbf{6.6} & \textbf{15.2} & 16.29 \\ 
		& \multirow{2}{*}{$\lambda=1$} & Weighted & 14.5 & 23.8 & 38.4 & 9.5 & 3.4 & 8.4 & 28.4 & \textbf{18.6} & \textbf{17.6} & 17.4 & 7.3 & 15.4 & 16.89 \\ 
		& & Top1 & \textbf{13.4} & 22.8 & 36.8 & 10.2 & 3.4 & 8.5 & 30.2 & 19.2 & 18.7 & 15.8 & 6.7 & 15.4 & 16.74 \\ 
		\hline
		\hline
		\multirow{4}{*}{cat-acc} & \multirow{2}{*}{$\lambda=0.1$} & Weighted & 0.94 & 0.89 & 0.95 & \textbf{0.98} & \textbf{0.96} & 0.94 & 0.83 & 0.67 & 0.93 & 0.78 & 0.94 & 0.93 & \textbf{0.8944} \\
		& & Top1 & \textbf{0.97} & 0.87 & 0.94 & 0.96 & \textbf{0.96} & \textbf{0.95} & \textbf{0.84} & 0.62 & \textbf{0.96} & 0.78 & 0.94 & 0.93 & 0.8930 \\ 
		& \multirow{2}{*}{$\lambda=1$} & Weighted & 0.92 & \textbf{0.91} & \textbf{0.96} & 0.91 & 0.94 & 0.82 & 0.80 & 0.70 & 0.85 & 0.75 & 0.91 & 0.93 & 0.8678 \\ 
		& & Top1 & 0.92 & 0.90 & 0.93 & 0.88 & 0.95 & 0.85 & 0.73 & \textbf{0.71} & 0.91 & \textbf{0.87} & \textbf{0.95} & \textbf{0.94} & 0.8775 \\ 
		\hline
	\end{tabular}
	\vspace{-5mm}
\end{table}

\myparagraph{Top-k category labels:} We also analyze performance when instead of the most-likely (top1) category label, the category network returns multiple labels (top2/3). To compute the pose error with multiple predicted category labels, we compute the viewpoint error with the pose output of every predicted category and take the minimum value.  For example, the pose error for the top3 category labels ($c_1, c_2, c_3$) using the notation of Sec.~\ref{sec:loss} is given by $\mathcal{L}_p(y^*, y(c_1, c_2, c_3)) = \min_{i=1..3} \mathcal{L}_p(y^*, y_{c_i})$.

As can be seen in Table~\ref{table:joint_top}, increasing the number of possible category labels leads to an increase in both category estimation accuracy and reduction in pose estimation error. However, it must also be noted that this reduction of pose error is very likely an artifact of the above metric ($\min_{i=1..3} \mathcal{L}_p(y^*, y_{c_i}) \le \min_{i=1,2} \mathcal{L}_p(y^*, y_{c_i}) \le \mathcal{L}_p(y^*, y_{c_1})$) because when we use an oracle for category estimation (GT), the viewpoint error is higher than top-2/3 error. At the same time, improving category estimation accuracy (comparing top1 and GT, $89.30 \rightarrow 100$) leads to better performance in pose estimation ($16.29 \rightarrow 15.28$).

\begin{table}
	\vspace{-8mm}
	\caption{Object category estimation accuracy and median viewpoint error when Top1/2/3 predicted labels are returned by the category network.}
	\label{table:joint_top}
	\centering
	\begin{tabular}{|cc|cc|cc|cc|}
		\hline
		\multicolumn{2}{|c|}{Top1} & \multicolumn{2}{|c|}{Top-2} & \multicolumn{2}{|c|}{Top-3} & \multicolumn{2}{|c|}{GT} \\
		cat-acc & pose-err & cat-acc & pose-err & cat-acc & pose-err & cat-acc & pose-err \\
		\hline
		0.8930 & 16.29 & 0.9455 & 14.08 & 0.9595 & 13.15 & 100 & 15.28 \\
		\hline
	\end{tabular}
	\vspace{-10mm}
\end{table}

\section{Conclusion and Future Work}
We have designed a new integrated network architecture consisting of a shared feature network, a categorization network and a new category dependent pose network (per-category collection of fully connected pose networks) for the task of joint object category and 3D pose estimation. We have developed two ways of fusing the outputs of individual pose networks and the output of the category network to predict 3D pose of an object in the image when its category label is not known. We have proposed a training algorithm to solve our joint network with suitable pose and category loss functions. Finally, we have shown state-of-the-art results on the PASCAL3D+ dataset for the joint task and have shown pose estimation performance comparable to state-of-the-art methods that solve the simpler task of 3D pose with known object category. 

\myparagraph{Future work} We are exploring two avenues of future research. (1) We have used a pose regression formulation for our fusion techniques but they can be extended to pose classification problems and a natural question is to ask how well our proposed network architecture performs for the joint task of pose label and category label estimation. (2) An extension of this work to the joint object detection and pose estimation task by incorporating the category dependent pose network into existing architectures. 

\myparagraph{Acknowledgement} This research was supported by NSF grant 1527340.

\bibliographystyle{splncs}
\bibliography{recognition,vidal,geometry,learning}

\appendix
\section{t-SNE Visualizations}
\label{sec:tsne}

t-SNE \cite{Maaten:JMLR08,Maaten:JMLR14} is a very popular algorithm to visualize feature representations learned using deep networks. We extract features at the end of the feature network from the following three networks: (1) Pre-trained ResNet-50, (2) Joint-weighted and (3) Joint-top1.
The first network is trained for the task of image classification on the ImageNet \cite{ImageNet} dataset. The second and third networks are the ones trained for our task of joint object category and pose estimation. Joint-weighted is the network trained with the weighted fusion strategy and Joint-top1 is the network trained with the top1 fusion strategy. For all three networks, we extract features at the end of Stage-4 (end of the feature network) and run the Barns-Hut implementation of the t-SNE algorithm \cite{Maaten:JMLR14}. 

\begin{figure}
	\centering
	\subcaptionbox{Features from Stage-4 of the ResNet-50 network. \label{fig:tsne_pose_car_base}}{\includegraphics[width=0.31\linewidth]{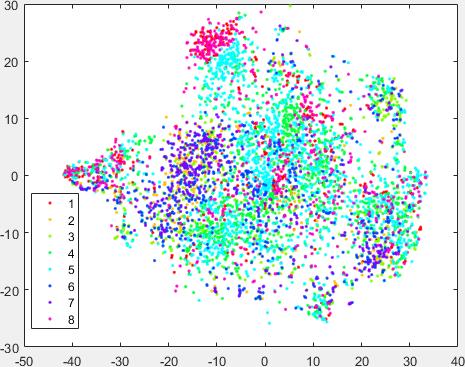}}
	~
	\subcaptionbox{Features learned by the Joint-weighted network.  \label{fig:tsne_pose_car_joint_wgt}}{\includegraphics[width=0.31\linewidth]{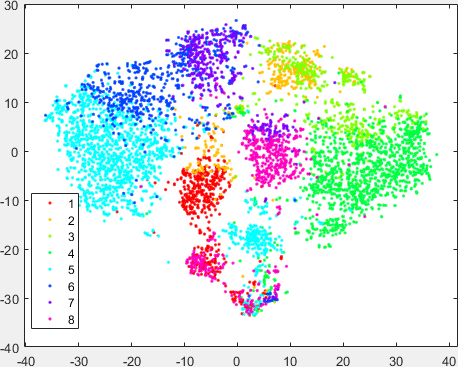}}
	~
	\subcaptionbox{Features learned by the Joint-top1 network. \label{fig:tsne_pose_car_joint_top1}}{\includegraphics[width=0.31\linewidth]{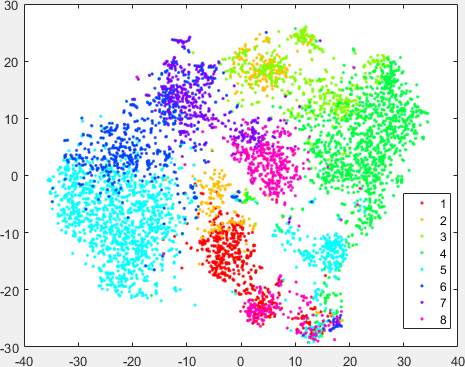}}
	\vspace{-1em}
	\caption{t-SNE visualization of car image features extracted from different networks with azimuth labels. The 8 azimuth labels were obtained by discretizing azimuth angle of rotation matrices into 8 bins.}
	\label{fig:tsne_pose_car}
	\vspace{-5mm}
\end{figure}

In Fig.~\ref{fig:tsne_pose_car}, we visualize the features from all car images with their azimuth labels. These azimuth labels are obtained by discretizing the azimuth angles into 8 bins. As can be seen in Fig.~\ref{fig:tsne_pose_car_base}, the pre-trained network has learned features that are not suitable for pose estimation. However, our networks trained for the joint category and pose estimation task under both fusion strategies (Figs.~\ref{fig:tsne_pose_car_joint_wgt}, \ref{fig:tsne_pose_car_joint_top1}) learn features that group cars in different orientations into different clusters.

\begin{figure}
	\centering
	\subcaptionbox{Features from Stage-4 of the ResNet-50 network. \label{fig:tsne_category_base}}{\includegraphics[width=0.31\linewidth]{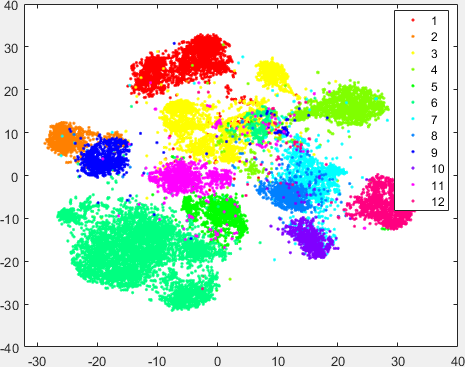}}
	~
	\subcaptionbox{Features learned by the Joint-weighted network. \label{fig:tsne_category_joint_wgt}}{\includegraphics[width=0.31\linewidth]{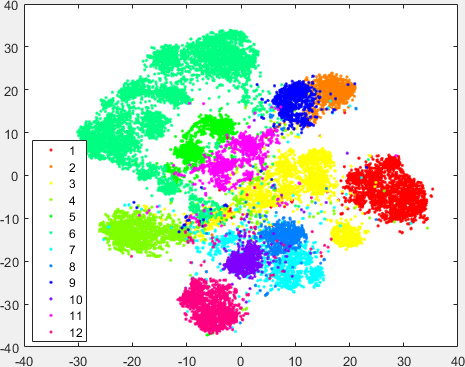}}
	~
	\subcaptionbox{Features learned by the Joint-top1 network. \label{fig:tsne_category_joint_top1}}{\includegraphics[width=0.31\linewidth]{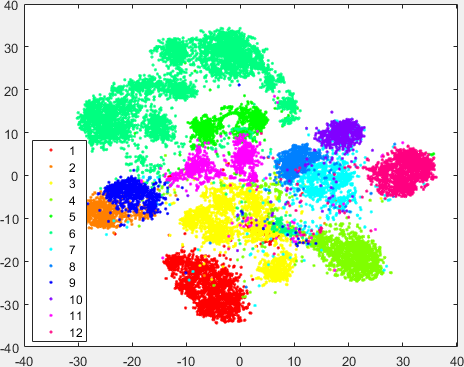}}
	\vspace{-1em}
	\caption{t-SNE visualization of features for all real images extracted from different networks with 12 object category labels.}
	\label{fig:tsne_category}
	\vspace{-8mm}
\end{figure}

In Fig.~\ref{fig:tsne_category}, we visualize the features from all real images across the twelve object categories with their object category labels. The image features of different object categories cluster into different groups for all three networks. This is to be expected in the pre-trained (Fig.~\ref{fig:tsne_category_base}) case but this is observed even in our joint networks (Figs.~\ref{fig:tsne_category_joint_wgt}, \ref{fig:tsne_category_joint_top1}).

\end{document}